\documentclass[letterpaper]{article}
\usepackage{flairs}
\usepackage{times}
\usepackage{helvet}
\usepackage{courier}
\usepackage{amsmath,amssymb,amsfonts}
\usepackage{algorithmic}
\usepackage{graphicx}
\usepackage{textcomp}
\usepackage{xcolor}
\usepackage{url}
\usepackage{xurl}
\begin{document}
%
\title{A Convolutional-based Model for Early Prediction of Alzheimer’s based on the Dementia Stage in the MRI Brain Images}
\author{Shrish Pellakur, Nelly Elsayed, Zag ElSayed, Murat Ozer\\
School of Information Technology\\
University of Cincinnati\\
}
\maketitle
\begin{abstract}
\begin{quote}
Alzheimer's disease is a degenerative brain disease. Being the primary cause of Dementia in adults and progressively destroys brain memory. Though Alzheimer's disease does not have a cure currently, diagnosing it at an earlier stage will help reduce the severity of the disease. Thus, early diagnosis of Alzheimer's could help to reduce or stop the disease from progressing. In this paper, we proposed a deep convolutional neural network-based model for learning model using to determine the stage of Dementia in adults based on the Magnetic Resonance Imaging  (MRI) images to detect the early onset of Alzheimer's.
\end{quote}
\end{abstract}

\noindent  Alzheimer's is a progressive brain disease that destroys the brain's memory and other mental functions~\cite{scheltens2016alzheimer,scheltens2021alzheimer}. It also causes Dementia in adults, a general term used to describe memory loss. It is the most common cause of Alzheimer's in adults. Dementia is not a normal part of aging~\cite{geldmacher1996evaluation}. It is caused by damage to brain cells and their connections. Alzheimer's is the most common cause
of Dementia. Nearly 60-70\% of Dementia cases are caused due to Alzheimer's~\cite {jindal2014alzheimer}. Alzheimer's is a specific disease, while Dementia is a generalized term used to describe memory loss or reduced mental ability to perform day-to-day activities.

Researchers believe that several factors may cause Alzheimer’s disease. These factors include genes, environment, and lifestyle. Aging is the primary known cause of Alzheimer’s disease, and the risk of developing Alzheimer’s disease increases with age. The risk of Alzheimer’s doubles every five years after age 65, which is why most people with Alzheimer’s are 65 or older. However, old age is not the only factor, as nearly 200,000 Americans have Alzheimer’s before age 65.
Genetics also play a risk factor for developing Alzheimer’s disease if the gene for Alzheimer’s runs in the family tree~\cite{bertram2012genetics}. When a disease is hereditary, it can be caused by environmental factors as well~\cite{armstrong2013causes}.

Brain disease diagnostics are mainly performed using either the recording of the brain signal using Electroencephalography (EEG) devices~\cite{babiloni2016brain},~\cite{zaghloul2015implementable},~\cite{pascual2011characterizing},~\cite{zaghloul2019early} or imaging techniques such as Magnetic Resonance Imaging (MRI) and Computed Axial Tomography (CT) scans~\cite{santavuori1989muscle,lemay1986ct,van1996wilson,imabayashi2013comparison,talo2019convolutional}. The Magnetic Resonance Imaging (MRI) scans technique is a non-invasive, painless, and safe test to obtain brain imaging. It uses waves (radio) and a magnetic field to produce detailed brain images. The process is similar to a CT scan, but it does not use radiation. The test can identify multiple physical damages caused to the brain in case of any external injury or stroke. Different stages of Dementia can be identified using Magnetic Resonance Imaging (MRI) scans. Using Deep Learning techniques, we can use the MRI images to train a Deep Learning model and use it to make accurate predictions in identifying different stages of Dementia. Diagnosing this at an earlier stage is very helpful and could reduce a lot of time and resources.

Using brain MRI images to detect the different stages of Dementia in Adults to detect the early onset of Alzheimer's using a Deep Learning model is an asset to the medical diagnosis field. It can help the medical specialist to diagnose the stage of Dementia in adults to detect the early onset of Alzheimer's and can be done anywhere with limited resources and in a time-consuming manner. In this paper, we proposed a deep learning-based model for the early detection of Dementia and for identifying the stages of Dementia in adults. We used the MRI images dataset benchmark OASIS database~\cite{marcus2010open}. The database includes MRI images of moderate, mild, very mild, and non-Dementia cases in adults. Mild and Very Mild Dementia describes the initial stages of Dementia. Moderate Dementia refers to a slightly more advanced stage, and non-Demented would be a stage before the Dementia essentially starts. In addition, we applied data augmentation to enhance the model's capability of detecting Dementia stages from different MRI-obtained images with different resolutions. Thus, our proposed model can successfully process different MRI brain images without restrictions on the image-capturing source resolutions.

\begin{figure*}[htbp]
	\centerline{\includegraphics[width=\linewidth,height=18cm]{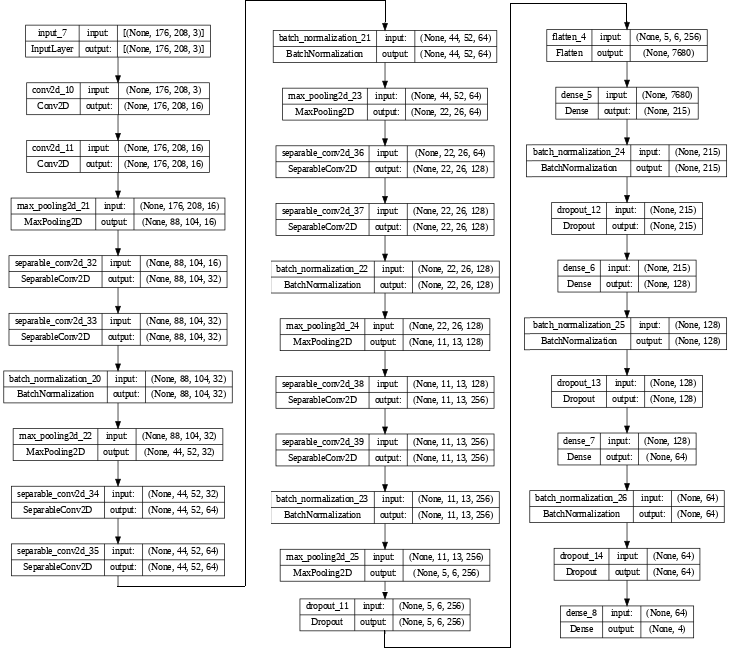}}
	\caption{The proposed model architecture diagram.}
	\label{mode_diagram}
\end{figure*}

\section{Related work}

Several computing-based brain disease diagnostic models have been designed using MRI imagins as the data sources. Sultan et al.~\cite{sultan2019multi} used brain MRI images to classify brain tumors. Sultan et al.~\cite{sultan2019multi} model was designed using a convolutional neural network that consists of  16 layers. The primary issue the model faced was overfitting due to the number of data sample limitations in the dataset. Demirhan et al.~\cite{demirhan2016classification} proposed a method using Machine Learning algorithms and techniques to diagnose Mild Cognitive Impairment (MCI) in adults, which will further help predict
the chances of Alzheimer's in that adult. MCI is a stage before Alzheimer's disease, which is not a part of healthy aging. Approximately 12\% of adults with MCI are likely to develop Alzheimer's disease within a year, and 80\% of the adults are likely to develop it within six years. In this research, MRI imaging is used as the MRI is sensitive to the degeneration which Alzheimer's disease causes on the brain, such as tissue damage or tissue loss.

Magnin et al.~\cite{magnin2009support} used histogram analysis for extracting features for Alzheimer's disease classification from whole-brain anatomical MRI. The model is based on the concept that Grey matter, White matter, Cerebral Spinal Fluid, and other brain tissues give out
information about neurodegenerative diseases like Alzheimer's using the support vector machine (SVM). Islam et al.~\cite{islam2018brain} developed a machine learning approach to classify the various stages of Alzheimer's based on the dementia level from brain MRI images using an ensemble system. In addition, the model also had three feature extractors prior to the model training. This approach, since it deals with small patches of the brain regions, can pick up the subtle changes in the brain and therefore helps find patterns of brain abnormalities.

The primary challenges faced in all the mentioned literature are model overfitting due to the lack of MRI datasets availability that addresses adults with Dementia or Alzheimer's disease, the accessibility involved in getting these images, and the high computational power required
to implement detection models. Some researchers have used transfer learning where previously built models have been used to solve a similar problem to overcome the resources restrictions challenge~\cite{ghazal2022alzheimer,ashraf2021deep,aderghal2018classification}. However, the dataset availability and quantity limitation is still the major limitation for building a robust Dementia or Alzheimer's disease detection model.

This paper tackles the data limitations by increasing the size of the dataset and adjusting the model to be compatible with different MRI image resolution sources. We proposed a novel deep convoluted neural networks-based model for detecting the Dementia stages from MRI brain images. In addition, we performed the data augmentation on the benchmark dataset using five functions to increase the dataset's size and the detection problem's complexity. Thus, the model can use different MRI image sources without compatibility restrictions due to image capturing resolution. Moreover, the applied data augmentation solved the model overfitting problem.

\section{Proposed Model}

The proposed Dementia stages detection model consists of five Convolutional blocks, which contain a combination of 2D convolutional layers, max-pooling layers, separable convolution layers, and batch normalization. The separable convolutions first perform a depthwise spatial convolution that acts on each input channel separately, followed by a pointwise convolution that mixes the resulting output channels~\cite{chollet2015keras}. Then the Flatten layer is followed by four fully connected layers. The rectified linear unit (ReLU) function has been used as the model activation~\cite{agarap2018deep,elsayed2019effects}. The proposed model diagram is shown in Figure~\ref{mode_diagram}.

\subsection{Data Augmentation}

In this study, five Data Augmentation functions have been used on the images:
\begin{itemize}
	\item\textbf{ Anti-Clockwise Rotation:} The images are rotated in an anticlockwise direction. The rotation degree rotation is set at a random number ranging from 0 to 180 degrees.
\item \textbf{Clockwise Rotation:} The images are rotated in a clockwise direction. The degree of rotation is set at a random number ranging from 0 to 180 degrees.
\item \textbf{Horizontal flip:} The images are flipped completely in a horizontal direction.
\item \textbf{Vertical flip:} The images are flipped completely in a vertical direction.
\item\textbf{ Blur-image:} It applies Gaussian blur to the image.
\item \textbf{Noise: }It adds random noise, either Gaussian or Uniform, to the images. This makes the used images contain real-world noises and enhances our model's robustness and performance to different image sources and resolutions.
\end{itemize}

After performing the data augmentation, the original dataset and the augmented images are merged into one dataset. Then, we split the
data into training, validation, and testing sets with the ratio 6:2:2, respectively. Then, image resizing was performed to set the height and the width of the image to 176 and 208 pixels, respectively.

\begin{figure}[htbp]
	\centerline{\includegraphics[width=\linewidth,height=4cm]{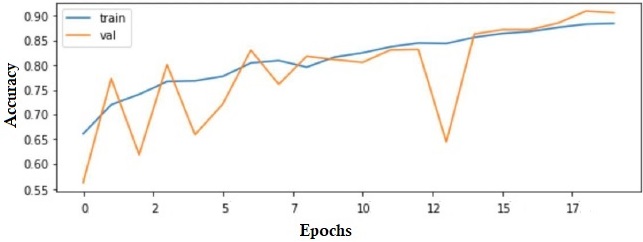}}
	\caption{The propsoed model training versus validation accuracies.}
	\label{auc_graph}
\end{figure}

\begin{figure}[htbp]
	\centerline{\includegraphics[width=\linewidth,height=4cm]{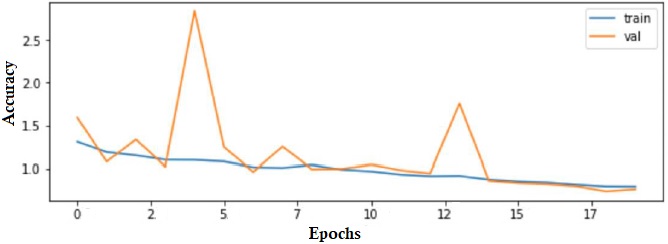}}
	\caption{The propsoed model training versus validation losses.}
	\label{loss_graph}
\end{figure}

\section{Results and Analysis}

The proposed model has been built and trained. The Adam optimizer has been used as the optimization function. The number of epochs was set to 20 epochs. The categorical cross entropy function has been set as the loss function. Figure~\ref{auc_graph} shows the training versus validation accuracies of the proposed model, and Figure~\ref{loss_graph} shows the training versus validation losses of the proposed model over 20 epochs.

Table~\ref{model_results} shows the proposed model's overall results for the training and testing stages. In addition to the number of parameters, including the trainable and non-trainable parameters.

\begin{table}
	\caption{The proposed model overall results and statistics.}
	\begin{center}
		\small
			\begin{tabular}{|l|c|}
				\hline
				\textbf{Metrics}&\textbf{Value}\\
				\hline
				Train Accuracy& 93.10\%\\
				Test Accuracy&90.05\%\\
				Train Loss&0.0690\\
				Test Loss& 0.0995\\
				\#Parameters&1,831,895\\
				Trainable Param.& 1,830,121\\
				Non-traibable Param.&1,774\\
				\hline
			\end{tabular}
		\label{model_results}
	\end{center}
\end{table}	

We compared our model to the current state-of-the-art Dementia stage detection models. Table~\ref{compare_table} shows the different models, the primary technique that has been used, and the overall accuracy of testing the model.

\begin{table}[t]
	\caption{A comparision between the proposed model and the state-of-the-art Dementia stage detection models.}
	\begin{center}
	\resizebox{\linewidth}{!}{	
			\begin{tabular}{|l|l|c|}
				\hline
				\textbf{Model}&\textbf{Teqchnique}&Accuracy\\
				\hline
				
				\cite{baglat2020multiple}&	Logistic Regression(w/imputation)  &  78.94\%\\
				\cite{baglat2020multiple}& Logistic Regression(w/dropping) &   75.00\%\\
				\cite{baglat2020multiple}&  	SVM&   81.57\%\\
				\cite{baglat2020multiple}& Decision Tree &   81.57\%\\
				\cite{baglat2020multiple}& 	Random Forest &   86.84\%\\
				\cite{basheer2021computational}&Capsule Netwok&80.38\%\\
				\cite{mahmood2013automatic}&PCA 150 Features&89.22\%\\
				\cite{mahmood2013automatic}&PCA 100 Features&86.47\%\\
				\cite{maqsood2019transfer}&Transfer Learning-ALexNet&89.6\%\\
				Our& CNN+Augmentation&90.05\%\\
				\hline
			\end{tabular}
			}
		\label{compare_table}
	\end{center}
\end{table}	
\section{Conclusion and Future Work}

Early detection of dementia is crucial to help patients receive the required medication to reduce the progressive damage of the brain and early detect Alzheimer's, especially as it does not have a current known treatment. In this paper, we addressed the dementia detection system challenges due to the data availability limitations by increasing the size of the dataset using data augmentation using five functions to increase the dataset's size and the detection problem's complexity. In addition, we proposed a novel deep convoluted neural networks-based model for detecting the Dementia stages from MRI brain images. In addition, it supported the model to overcome the overfitting problem. Thus, the model can use different MRI image sources without compatibility restrictions due to source resolutions. Furthermore, the proposed model shows a significantly higher accuracy compared to the state-of-the-art architectures with a convenient implementation budget.

\bibliography{deepImage_references}
\bibliographystyle{flairs}

\end{document}